# Actigraphy-based Sleep/Wake Pattern Detection using Convolutional Neural Networks

Extended Abstract

Lena Granovsky
Analytics and Big Data, Teva R&D
PO Box 8077, Israel
lena.granovsky@teva.co.il

Gabi Shalev
Analytics and Big Data, Teva R&D
PO Box 8077, Israel
gabi.shalev02@teva.co.il

Nancy Yacovzada
Analytics and Big Data, Teva R&D
PO Box 8077, Israel
nancy.yacovzada@teva.co.il

Yotam Frank
Analytics and Big Data, Teva R&D
PO Box 8077, Israel
yotam.frank01@teva.co.il

Shai Fine
Analytics and Big Data, Teva R&D
PO Box 8077, Israel
shai.fine01@teva.co.il

## ABSTRACT

Common medical conditions are often associated with sleep abnormalities. Patients with medical disorders often suffer from poor sleep quality compared to healthy individuals, which in turn may worsen the symptoms of the disorder. Accurate detection of sleep/wake patterns is important in developing personalized digital markers, which can be used for objective measurements and efficient disease management.

Big Data technologies and advanced analytics methods hold the promise to revolutionize clinical research processes, enabling the effective blending of digital data into clinical trials. Actigraphy, a non-invasive activity monitoring method is heavily used to detect and evaluate activities and movement disorders, and assess sleep/wake behavior. In order to study the connection between sleep/wake patterns and a cluster headache disorder, activity data was collected using a wearable device in the course of a clinical trial.

This study presents two novel modeling schemes that utilize Deep Convolutional Neural Networks (CNN) to identify sleep/wake states. The proposed methods are a sequential CNN, reminiscent of the bi-directional CNN for slot filling, and a Multi-Task Learning (MTL) based model. Furthermore, we expand standard "Sleep" and "Wake" activity states space by adding the "Falling asleep" and "Siesta" states. We show that the proposed methods provide promising results in accurate detection of the expanded sleep/wake states. Finally, we explore the relations between the detected sleep/wake patterns and onset of cluster headache attacks, and present preliminary observations.

## 1 INTRODUCTION

Sleep/wake patterns and sleep disturbance are important in many health-related areas for diagnostic purposes and disease management. Assessing sleep quality and daytime wakefulness potentially allows more effective disease management and treatment, and offers a chance for better understanding of sleep- and wake-promoting mechanisms.

Patients with chronic medical disorders often have fewer hours of sleep and less restorative sleep compared to healthy individuals, and this poor sleep may worsen the symptoms of the disorder [1]. As an example, patients suffering from Alzheimer's disease, exhibit significant changes in sleep/wake patterns quite early in the disease process. Disruptions of nighttime sleep increase in magnitude with increasing severity of dementia [2]. Sleep disruptions are also related to control of respiratory diseases such as asthma and COPD. Not well-controlled asthmatic patients tend to awake more, take longer to fall asleep, and spend more time awake during the night compared to those with well-controlled asthma [3]. For reasons, which are partly unknown, headache and sleep share an especially close relationship [4]. The interaction between sleep and headache is known to be complex, and evidence suggests that it may be of critical importance in our understanding of primary headache disorders, such as migraine and Cluster Headache [5].

Several methods exist to collect sleep/wake data. Polysomnography test, in which multiple channels of sleep data are collected in a laboratory, has long been the "gold standard" for sleep assessment [6]. However, the gathering of polysomnographic data is invasive, costly, disruptive to sleep, and is typically performed for only one or two nights.

Actigraphy has been used for over 25 years to assess sleep/wake behavior [7]. It is a non-invasive activity monitoring device (wearable sensor), usually worn on a wrist, which allows continually measuring motor activity for long time periods. There are many benefits of utilizing actigraphy in medical settings: it is less cumbersome than the polysomnography test, less expensive, and more reliable as it can be worn for extended periods of time [7]. The data can be either downloaded or streamed to a computerized backend platform and analyzed, either offline or in real time (depends on the target use-case and the nature of the application). These properties make it potentially useful for gathering objective motor activity in large population studies, in which issues of participant burden are important.

A typical way of modeling activity data in clinical settings usually consists of three steps. The descriptive analytics step uses signal processing methods to analyze raw activity time-series data (both at the time and the frequency domain). Then, the diagnostic analytics step detects activity states and movement patterns derived from the raw activity data. The states can be either fixed (like "Sit" or "Sleep"), or transition (like "Falling asleep" or "Passaging from sitting to standing"). Finally, in the prognosis and prescriptive analytics step digital markers and predictive models are developed to detect, monitor, and predict a certain medical condition.

The focus of this study is on the diagnostic analytics step, where the goal is to reliably detect and monitor sleep/wake patterns. Accurate detection of "Sleep" and "Wake" states, as well as detection of additional states like "Falling asleep" and "Resting", can be crucial in discovering personalized digital markers used for disease management and prediction of disease deterioration.

Detecting sleep/wake states using actigraphy data is the subject of many efforts [8][9][10][11]. In a paper from 2001, Jean-Louis et al. [9] applied a sleep/wake detection algorithm to the 24h recordings from accelerometers worn by two separate validation study groups (post-menopausal women and young adults), and achieved a minute by minute agreement rate of 85%-91%. In another study, that quantified wrist movements data, Jean-Louis et al. [10] managed to achieve an even higher agreement rate (91.4%-96.5%), while showing correlation of 0.79-0.94 with sleep duration. In a paper from 2004, Hedner et al. [11] developed an adaptive algorithm for sleep/wake assessment in sleep apnea patients from measurements of their wrist actigraphy. Their algorithm achieved overall sensitivity of 89% and 69%, respectively, in the task of identifying sleep, and the agreement ranged from 86% in the normal subjects to 86%, 84%, and 80% in the patients with mild, moderate, and severe obstructive sleep apnea (respectively).

Notwithstanding the above successes, there are several challenges in achieving a reliable sleep/wake detection using the traditional methods. For example, quantifying activity counts on 30-seconds epoch basis, as usually done in actigraphy, may lead to an underestimation of wake periods where the subject shows reduced body movements. This is particularly the case for subjects with insomnia symptoms who often have a large proportion of wake epochs during overnight sleep. Figure 1, adapted from Long et al. [8], provides an interesting perspective over the correspondence between the activity level and identifiability of sleep and wake epochs. The figure presents a normalized histogram of sleep (light gray bars) and wake (dark gray bars) epochs binned into 7 buckets, where the leftmost bucket represents very low activity, and rightmost bucket represents high activity. While one would expect a negligible percentage of wake epochs at the low activity buckets, surprisingly, the highest percentage of wake epochs appears at the '0-5' actigraphy bucket. The reverse phenomenon is less dominant but still apparent – sleep epochs are recorded in moderate to high activity levels (actigraphy 10 and above).

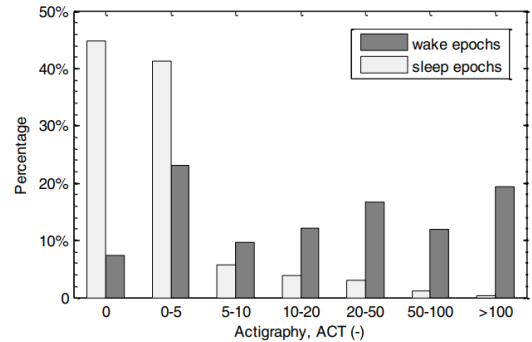

**Figure 1: Normalized histogram of epoch-based actigraphy data (ACT) in different levels of activity. Source: "*Actigraphy-based sleep/wake detection for insomniacs*", by Long et al. [8].**

In their paper, Long et al. provide the following explanation for this phenomenon, highlighting some of the challenges associated with modeling sleep/wake states using actigraphy data:

- Subjects with insomnia symptoms have a large proportion of wake epochs during overnight sleep
- Over 40% of wake epochs are of relatively low activity and resemble sleep epochs
- Restless sleep epochs (with body movements or arousals) are difficult to distinguish from wake epochs

Similar difficulties were also witnessed with the data used in our study. Figure 2 presents an example of activity data collected from a single patient during a period of 30 hours. The data is manually annotated into two periods of "Wake" surrounding the "Sleep" period. The blue line, marked "Enhanced Annotation", provides a finer grain segmentation, in which one can distinguish between two activity levels during the "Wake" period.

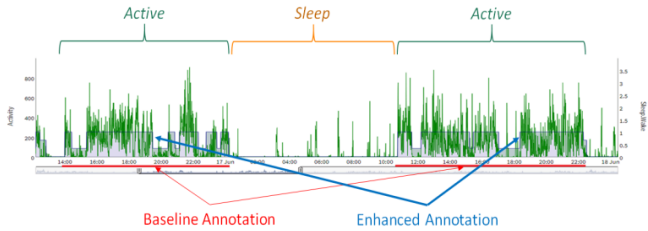

**Figure 2: Sample of baseline and enhanced annotations of the actigraphy activity data collected by a wrist-worn device, from a single patient during a period of 30 hours.**

These observations, together with the insights provided by Long *et al.*, have led us to suggest an expansion of the two typically used states ("Sleep" and "Wake") by two additional states - "Falling asleep" and "Siesta":

- "Falling asleep" (cf. Figure 3) - defined by a low density or a lower activity level, comparing to the "Wake" state. In addition, "Falling asleep" state must be preceded by a "Wake" state and followed by a "Sleep" state (context depended). Thus, "Falling asleep" is an example of a transition state.
- "Siesta" (cf. Figure 4) – defined by a very low activity occurred *only* during day-time. It represents either napping or just resting, and it is surrounded by "Wake" states. Hence, unlike "Sleep", there is no "Falling asleep" transition state leading to a "Siesta" state.

Over the last few years, deep neural networks (NNs) have achieved state of the art performance on a wide variety of machine learning tasks in various domains such as computer vision, speech recognition and natural language processing. Recently, there has been a lot of interest in adapting NNs to human activity recognition [12] and sleep/wake state detection [13].

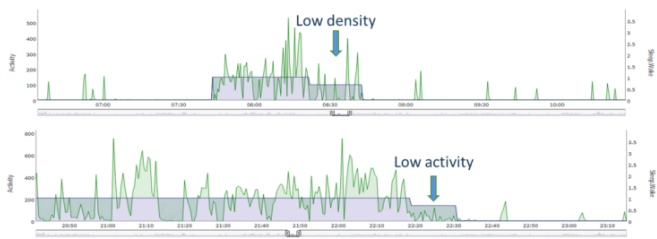

**Figure 3: Description of the "Falling asleep" state based on actigraphy data. In the upper chart, the "Falling asleep" state is determined by lower density comparing to the "Wake" state, while in the lower chart the "Falling asleep" is determined by lower activity levels.**

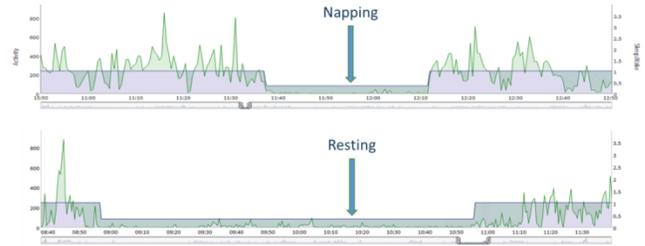

**Figure 4: Description of the "Siesta" state based on actigraphy data. The "Siesta" state is determined by a very low activity levels, and is surrounded by higher activity levels representing "Wake" states.**

This study presents two methods that utilize a 1-dimensional Deep Convolutional Neural Networks (CNN), designed to automate sleep/wake patterns detection from raw actigraphy motor sensor data. The first method is a sequential CNN, reminiscent of the bi-directional sequential CNN for slot filling. The second method is based on a Multi-Task Learning (MTL) approach.

The proposed models were trained and validated using the motor actigraph data collected from cluster headache patients in the course of a clinical study. They compared favorably to the performance obtained by two forms of Recurrent Neural Networks (RNN), i.e. Long Short Term Memory (LSTM) and Gated Recurrent Unit (GRU), and also to the performance obtained by a standard multilayer perceptron (5 hidden layers) that utilized a set of augmented (manually engineered) features.

Finally, we explore the relations between sleep/wake patterns, as detected by the proposed algorithms, and the onset of cluster headache attacks, and present preliminary observations. This will enable, in the prognosis and prescriptive analytics step, to develop an end-to-end mobile health (mHealth) solution for detecting and predicting the onset of cluster headache attacks.

## 2    DIGITAL CLINICAL STUDY SETTINGS

Cluster Headache (CH) is a primary headache syndrome characterized by strictly unilateral severe pain with accompanying autonomic symptoms. CH attacks often awaken a patient in the middle of the night with intense pain, and are accompanied by agitation and restlessness [4]. Although sleep and the cluster headache attacks are believed to be interconnected, the precise nature of the relationship between them remains obscure [14].

The data forming the basis of analyses in this paper was collected from 25 chronic CH patients participating in a clinical study. The objective of the study was to evaluate how the changes in sleep and wake parameters and physical movement relate to the occurrence of CH attacks. During 12 experimental weeks, the subjects were asked to continuously wear a wrist-watch-type actigraph device on the non-dominant hand. The device, which recorded arm movement, assessed the motor activity by calculating the Euclidian norm of the three-axis accelerometer data collected at 30Hz frequency. The collected data was then transformed to an aggregated activity measure, computed every 30-second period.

In addition to monitoring the activity data, subjects were equipped with an electronic diary device to record a daily report of the frequency, duration and intensity of CH attacks. The electronic diary reports were then matched with the activity data in order to assign a binary yes/no CH attack tag for every 30-seconds period (epoch).

The data collected by the actigraph device, as well as the reports from the electronic diary, were transferred to a cloud-based digital health platform solution for further analysis, modeling, and scoring (see Figure 5). This is an example for a typical mobile health (mHealth) system design, utilizing Internet of Things (IoT) and Big Data technologies. It holds the promise to revolutionize clinical research processes, enabling a seamless and effective blending of real-world data into clinical trials. The use of IoT in clinical trial makes the process more efficient and cost-effective, reducing the time needed to research new treatments.

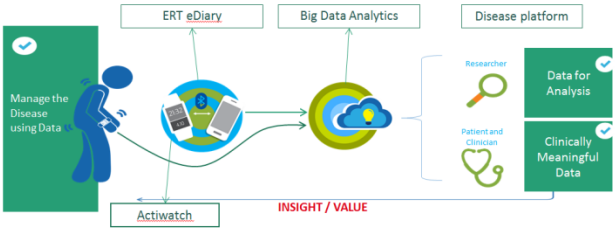

Figure 5: Cloud-based digital health platform in the Cluster Headache clinical study. This is an example for a mHealth system design, utilizing IoT and Big Data technologies.

## 3 METHODS

In this Section we describe two methods of deep CNN for an automated identification of sleep/wake states from raw motor activity data.

### 3.1 Data Representation

The data representation used in this study is based on a 721 dimensional feature vector, extracted by a sliding window over a 1-Dim activity level time series (per patient), where each point is a (smoothed, down-sampled) 30 seconds activity level. Thus, the 721-Dim feature vector represents an activity point of interest surrounded by 6 hours "context", i.e. 3 hours of activity recordings before and 3 hours after the point of interest. The data is labeled with four ordinal labels: "Wake", "Falling asleep", "Siesta", "Sleep" (cf. Figure 6).

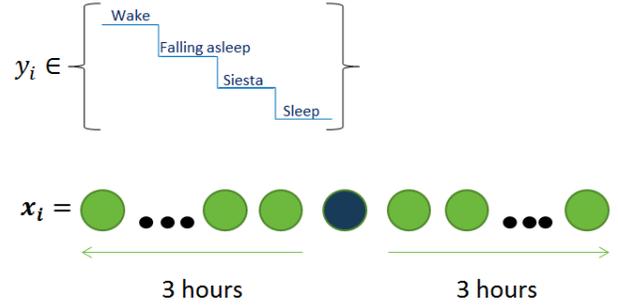

Figure 6: Description of the data used for modeling: four ordinal labels, and a feature vector surrounded by the three hours of context before and after the point of interest.

State detection within a context can be considered as a task of extracting local motifs. To this end, an attractive modelling approach is deep convolutional neural network (ConvNet, CNN). Convolutional networks are rooted in ideas from "NeoCognitron" [15] and inspired by Hubel's and Wiesel's work on visual primary cortex [16]. Their first modern form appeared in LeCun's work to recognize handwritten digits (LeNet) [17] Convolutional networks reached their current status as visual recognition de-facto standard, after beating traditional computer-vision techniques in ILSVRC 2012 competition by huge margin (AlexNet) [18].

A leading premise of ConvNets is that shared properties of the data can be described by a reduced set of parameters that needs to be learned. Thus, learned features at one part can be applied to other parts of the data. The architecture of a typical ConvNet is structured as series of stages, where

- Convolutional layers – detect local conjunctions of features from the previous layers
- Pooling layers – reduce representation size and select most informative features, thus reducing the number of parameters and computation in the network, controlling model complexity and overfitting

Convolutional layers consist of a set of trainable weight kernels that produce their output values by cross-correlating with input data. Thus, in computer vision tasks, every spatial patch in the image is weighted using the same shared kernels. In each layer, for every spatial location there are multiple values describing it. The different values for each location are known as the layer's feature maps. Denote the learnable kernel weights as $w$, the input as $r_{i,j,k}$ and output as $z_{i,j,k}$, where $i,j$ is the spatial location and $k$ is the corresponding feature map. The computed function for each feature map in a convolution layer $l$ is

$$z^l_{i,j,k} = \sum_{f=0}^{F-1}\sum_{m=0}^{M-1}\sum_{n=0}^{N-1} w_{n,m,f,k} r^{l-1}_{(i+n),(j+m),k} + b_k \quad (1)$$

Since convolutional layers contain weights that are shared by multiple spatial regions, they naturally aggregate and average the gradients over large amounts of data.

The same function can be applied as 1-Dim convolution on temporal data such as audio and text. In this case, the computed function for each feature map in a convolution layer $l$ is

$$z_{i,k}^l = \sum_{f=0}^{F-1}\sum_{n=0}^{N-1} w_{n,f,k} r_{(i+n),k}^{l-1} + b_k \qquad (2)$$

where $w$ is the kernel weights over the 1-Dim (sliding) window of the temporal data (e.g. activity levels).

The ConvNet architecture enables the detection of local motifs (extracted features), which in turn are invariant to location and scaling. For the task at hand, sleep/wake state detection using 1-Dim activity data, this architecture provides interplay between

- Convolution layer – Holds multiple filters; Each filter extracts a local time-dependent pattern within context (6h) boundaries
- Pooling layer – Reduces the output's dimensionality while keeping the most salient features and provides location and scaling invariance
- Inter context (feature vectors) dependency achieved via meta features created in the deeper network layers

Another natural approach for sleep/wake detection is to treat it as a sequence, i.e. time dependent pattern recognition, rather than a "static" (time invariant) state detection. Thus, sleep/wake detection may be considered as an instance of sequence labeling. The goal of sequence labeling task is to assign, for each item in the input sequence, a label drawn from a fixed label set. Over the last few years, Recurrent Neural Network (RNN) models have increasingly been used for various sequence labeling task, such as POS tagging [19], named entity recognition [20], and slot filling [21]. For those tasks, accuracy is generally improved by making the optimal label for a given element dependent on the choices of nearby elements. However, RNNs are difficult to train due to the vanishing and exploding gradient problems [22]. The exploding gradient problem can be avoided by using a simple strategy of gradient clipping [23]. Several architectures were proposed to deal with the vanishing gradient problem, such as LSTM [24] and GRU [25]. However, these methods increase the number of model parameters. It was shown that LSTM fails to learn to correctly process certain very long or continual time series, which are not a priori segmented into appropriate training subsequences with clearly defined beginnings and ends [26].

A recent study have investigated the usage of CNNs for a sequential labeling task (slot filling task) [21], and showed that it outperforms recurrent neural networks on this task. Inspired by this recent success, our modeling approach is designed to take an advantage of the reduced complexity of the CNN architecture (relative to RNN), while leveraging the inherent strength of CNN to detect local motifs in the augmented 6 hours context. Time dependency is achieved by extracting feature vectors using an overlapping sliding window (a common practice heavily used, e.g. in the speech realm).

In the sequel, we will present two 1-Dim CNN-based models. The first is a sequential CNN, reminiscent of the bi-directional sequential CNN for slot filling, and the second is resulted by employing a multi-task learning approach. We will conclude this section by comparing the results of the two suggested CNN-based models with the ones obtains with LSTM, GRU, and standard multilayer perceptron (5 hidden layers) using augmented (manually engineered) feature set.

### 3.2 Sequential CNN Model

A bi-directional sequential CNN (bi-sCNN) was suggested by Ngoc Thang Vu [21] to model a sequential labeling in spoken language understanding (SLU), a.k.a. the slot filling task. The goal is to provide a semantic tag for each of the words in a given sentence. For example in the sentence: "I want to fly from Munich to Rome" a SLU system should tag Munich as the "departure city" of a trip and Rome as the arrival city. All other words, which do not correspond to real slots, are tagged with an artificial class "O". In [21] it was shown that bi-sCNN method outperforms recurrent neural networks on this task.

We hereby give a short description of the bi-sCNN model. The architecture of the network is presented in Figure 7. Each sentence in the dataset was divided into past and future parts (with overlap), and the model was comprised of two sequential CNN "twin models": one for the past part of the sentence and the other for the future part (hence the bi-directionality). Given a sentence **s** and a center word $w_t$, whose label needs to be predicted, each of the two models were fed in with $2D$ image comprised of the concatenation of the distributed representations of the words in its respective part of **s**. The $T \times d$ image ($d$ is the dimensionality of the word representation and $T$ is the length of the sentence part) was then convolved with filters of size $f \times d$ (spanning all of the embedding dimensions) to obtain multi-channel 1-Dim representations of the original sentence. Following this first step, a max-pooling operation was applied to each of the channels followed by more convolution-pooling steps to obtain a representation vector $C_{p_t}$, $C_{f_t}$ of the input. The resulting representation was then combined (either by mathematical operations or by concatenation) with the distributed representation of the current word $e(w_t)$ to obtain a hidden $h(w_t)$ layer output that was used in turn to predict the semantic label of $w_t$.

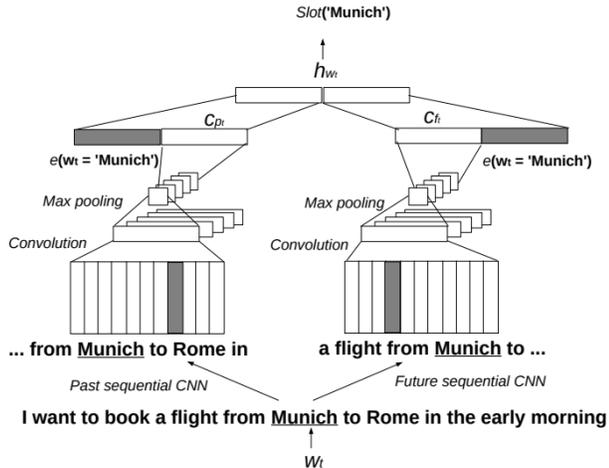

**Figure 7: Bi-directional sequential CNN that combines past and future sequential CNNs for slot filling. Source:** "*Sequential convolutional neural networks for slot filling in spoken language understanding*", by Vu, N. T [21].

Inspired by the successful application of the bi-sCNN model, we aimed to take advantage of the reduced complexity of CNN architecture (relative to RNN), while leveraging the inherent strength of CNN to detect local motifs in the time augmented context.

Our first modeling approach to the sleep/wake state detection differs in that, contrary to the above mentioned model, our architecture includes only one sequential CNN model and uses as input a 1-Dim vector of size 721 representing the activity measurement at a particular point in time and 3 hours before and after. Hence we are using a 6 hour time window. Also, given the 1-Dim input, all of the convolution layers that we use are also one dimensional. Moreover since our descriptor of the current time step is one number (the activity measurement in a certain point in time) we do not concatenate it to the following layers in the network, rather we use only the higher level representation vectors that we obtain by applying convolution and max pooling in order to predict the current state of the subject (i.e. whether he/she is awake or asleep).

We applied the sequential CNN model to the actigraphy data collected from cluster headache in the course of a clinical study. From each patient, a time series of 75K to 90K epochs was collected. This time series is subsequently divided into windows of an equal size. The data was divided into 80% / 20% train and test proportion. Table 1 and Table 2 present the performance evaluation of the model.

**Table 1: Sequential CNN performance by sleep/wake states**

|           | Sleep | Siesta | Falling asleep | Wake  |
|-----------|-------|--------|----------------|-------|
| Precision | 0.998 | 0.972  | 0.982          | 0.996 |
| Recall    | 0.998 | 0.984  | 0.981          | 0.994 |

**Table 2: Sequential CNN confusion matrix**

| Predicted/Actual | Sleep  | Siesta | Falling asleep | Wake   |
|------------------|--------|--------|----------------|--------|
| Sleep            | 36,596 | 1      | 29             | 27     |
| Siesta           | 5      | 7,684  | 8              | 203    |
| Falling asleep   | 18     | 4      | 3,711          | 45     |
| Wake             | 20     | 116    | 34             | 53,995 |

### 3.3 Multi-task CNN Model

Multi-task learning (MTL) is an approach to learn multiple related tasks simultaneously in such a way that knowledge obtained from each task can help improve the learning of other tasks. To this aim the various tasks are learned in parallel while using a shared representation. As suggested by Caruana [27], MTL may make learning faster, more accurate, and enable the learning of tasks which otherwise would have been too difficult to learn.

The common deep MTL architecture suggests incorporating supervision for all tasks at the same layer, which is the last layer of the network. This layer is shared among all the tasks. Recent work by Søgaard *et al*. [28] showed that having task supervision from all tasks at the same last layer is often suboptimal. Hence, they presented an MTL architecture in which the various tasks supervision are at different layers, having low level task supervision at a lower layer and other task supervision at the last layer of the network.

Our proposed MTL architecture is based on the sequential CNN for activity sequence tagging. In our network we have two task supervisions (cf. Figure 8):

1) Distribution over total sleep and total awake time – given a sequence of activities, we would like to score the proportion of the total sleep time, whether it is "Sleep" or "Siesta" state. For example, given a labeled sequence of size $T$, denote by $S$ the number of epochs which are of "Sleep" or "Siesta" states. Thus $1 - \frac{S}{T}$ is the probability of total awake states, namely the proportion of being at either "Wake" or "Falling asleep" state. This supervision happens at a lower layer (cf. Figure 8, left branch)
2) Refined (4 level) state detection – given a sequence of activities, we would like to accurately detect the correct refined state of the current activity epoch: "Wake", "Falling asleep", "Siesta", "Sleep" (cf. Figure 8, right branch)

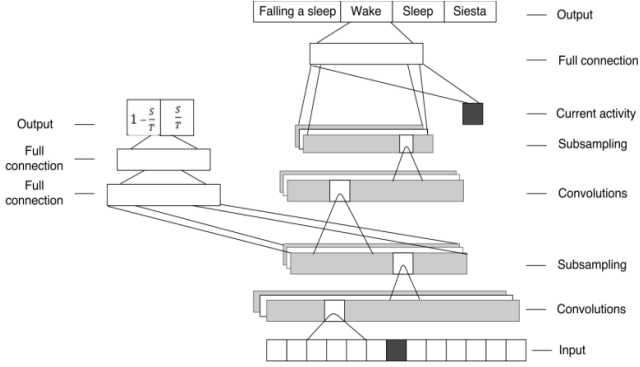

**Figure 8: Schematic representation of a Multi-Task CNN.**

A more detailed description of the proposed architecture is depicted in Figure 9: The main branch is formed by three stacked convolution layers followed by a max pooling layer after each convolutional layer. Then the intermediate representation is fed as an input to the bifurcated network. The left branch (marked (1)) includes a multilayer perceptron (MLP) with output layer of two neurons, which estimates the distribution over total sleeping time for the input segment. The right branch (marked (2)) stacks additional convolutional layers and fully connected layers, which is an intermediate representation of the entire sequence input. Before the last fully connected layer of branch (2) we concatenate the "current activity" epoch with the intermediate representation. This concatenation, as was demonstrated by [21], is used to help the model direct special attention to the ("current activity") epoch of interest which should be labeled.

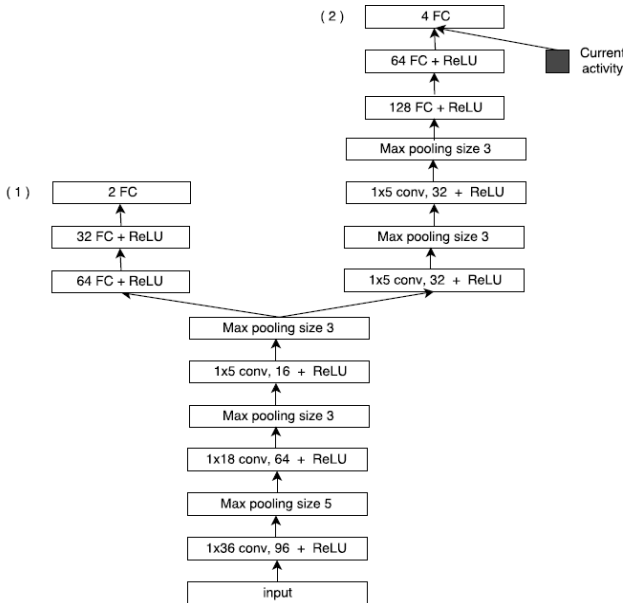

**Figure 9: Detailed architecture of Multi-Task CNN.**

We applied the Multi-Task CNN on the same data set that was used for training and testing the sequential CNN. Performance evaluation of the model is presented in Table 3 and Table 4.

**Table 3: Multi-Task CNN performance by sleep/wake states**

|           | Sleep | Siesta | Falling asleep | Wake  |
|-----------|-------|--------|----------------|-------|
| Precision | 0.998 | 0.974  | 0.982          | 0.998 |
| Recall    | 0.999 | 0.988  | 0.986          | 0.995 |

**Table 4: Multi-Task CNN confusion matrix**

| Predicted/Actual | Sleep  | Siesta | Falling asleep | Wake   |
|------------------|--------|--------|----------------|--------|
| Sleep            | 36,613 | 3      | 17             | 32     |
| Siesta           | 17     | 7,718  | 10             | 171    |
| Falling asleep   | 9      | 2      | 3,732          | 58     |
| Wake             | 0      | 82     | 23             | 54,009 |

While the Multi-Task CNN provides only moderate improvement over the sequential CNN, it converges faster and also it provides some interesting insights in the interplay between the two tasks. Apart from the interest in employing a multi task learning approach, we believe that this architecture will prove useful in handling more difficult detection settings.

## 3.4 Comparison with Additional Modelling Approaches

Finally, we compared the results of the two suggested CNN-based models with commonly used modeling approaches

- Standard Neural Network (MLP)
  - Data is augmented with 80 constructed features representing short and long term dependencies
  - 5 hidden layers, where number of neurons in each layer are: 1024, 512, 64, 128, 32
  - Transfer function RelU
  - We use batch normalization and dropout as regularization.
- Bi-directional LSTM
  - 128 units for the LSTM, and on top of it we stacked 2 fully connected layers with 512 neurons in the first layer and 32 neurons in the second
- CNN + Bi-directional gated recurrent unit (GRU)
  - The model includes 96 convolutional filters of size 1x16, and on top of it we added 32 convolutional filters of size 1x8, and then bi-directional GRU with 64 cells and 2 fully connected layers of 512 neurons and 32 neurons in the first and second layers, resp.

The CNN-based models were trained with stochastic gradient descent using a momentum of 0.9. We used mini-batches of size 32 and a cross-entropy loss function. For the multi-task CNN we used two cross-entropy loss functions, one per each task. The summation of the two was considered as the loss function of the model.

The comparison results, depicted in Figure 10, present a much faster convergence rate for the CNN-based models. The graphs also show a substantial improvement in accuracy over the recurrent architectures LSTM and GRU. Interestingly, a standard 5 layers, which includes as input the 80 constructed features, obtained great accuracy results. The advantage of the CNN based architecture is the fact that handcrafting features is unnecessary.

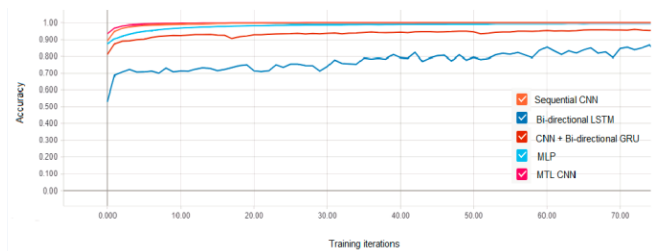

Figure 10: Comparison of the convergence rates of the two CNN-based models, vs LSTM, GRU and a standard MLP which uses an augmented feature set.

## 4 DISCUSSION

Although sleep patterns and medical conditions are known to be interconnected, the exact nature of the relationship remains obscured [5]. A better understanding of this relation may lead to more effective therapeutic regimes and prediction of disease deterioration. This work provides a tool that shed a light on sleep/wake characteristics and enables further research of the relationship between sleep/wake patterns, physical movement and medical indications.

We suggested a refinement to the standard "Sleep" and "Wake" state space by expanding it with two additional states: "Falling asleep" and "Siesta" (resting). Then, we described two methods of deep CNN for detecting the four sleep/wake states in a raw activity data: sequential CNN, and Multi-Task Learning (MTL) based approach. The proposed modeling approaches are designed to take advantage of the reduced complexity of CNN architecture (relative to RNN), while leveraging the inherent strength of CNN to detect local motifs in a time augmented context. Time dependency across adjacent feature vectors is preserved because of the use of an overlapping sliding window.

The proposed methods were evaluated on data collected from CH patients in the course of a clinical trial. The very high precision and recall rates, together with the results presented in the confusion matrices, indicate that the two methods accurately detect each one of the four sleep/wake states: "Sleep", "Wake", "Falling asleep" and "Siesta". The suggested CNN-based models were also compared with LSTM, GRU, and standard multilayer perceptron (with augmented manually engineered feature set), and they demonstrated a substantial improvement in accuracy and faster convergence rates.

The next step is to explore the relations between the sleep/wake patterns and the onset of CH attacks. This will allow researchers to identify meaningful patterns and changes in sleep/wake parameters before and during CH attacks. This form of analysis is an integral part of the prognosis and prescriptive analytics step, which upon success will enable the development of an end-to-end mobile health (mHealth) solution for detecting and predicting the onset of cluster headache attacks. Although this is still a work in progress, we present here a few preliminary observations and insights.

First, we applied the sequential CNN approach to automatically identify the four sleep/wake states from the raw activity data. The algorithm was applied on the actigraph data collected from the CH patients participating in the clinical study, and provided a time series of the detected sleep/wake states, which is in sync with the 30 seconds epochs. We then matched the patient's self-assessment reports of CH attacks (provided in the eDiary as a part of the study) with the four sleep/wake states, creating a binary "yes/no CH attack" label for every epoch. The full time series data was then partitioned into non-overlapping periods of 24 hours ("days"). Unweighted Pair Group Method with Arithmetic Mean (UPGMA) algorithm of hierarchical clustering was then applied on the daily sleep/wake time series data to check the hypothesis that days with CH attacks can be distinguished from the days without attacks. We use the Dynamic Time Warping (DTW) as a distance metric in the hierarchical clustering analysis [29]. DTW is known to be one of the most commonly used distance measures for comparison of time series data.

The clustering algorithm provides a dendrogram, in which days are represented by leaves, groups of days by inner nodes, and distances between days/day-groups by the length of the edges that connect them. Figure 11 depicts a dendrogram created by applying the clustering algorithm on the daily sleep/wake data of 16 consecutive days (from Aug 1$^{st}$ to Aug 18$^{th}$ of 2017). The data was extracted from two CH patients, who did not report any cluster headache attack during that time period. It is easy to see that no recognizable separation between the "attack free" days of the two subjects can be found.

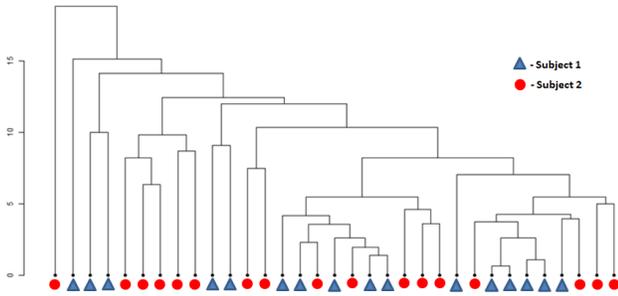

**Figure 11: Dendrogram produced by clustering "attack free" days (days w/o any CH attack) of two CH patients– no recognizable separation can be found.**

On the other hand, Figure 12 depicts a dendrogram of daily sleep/wake patterns data collected from two CH patients during 22 consecutive days (from Oct 26$^{th}$ to Nov 17$^{th}$ of 2017). During that time period, one of the patients experienced many cluster headache events (18 days out of 21), while another patient did not experience any CH attack. In this case, the clustering produces a perfect separation between the days of the patient suffering from CH attack and days of "attack free" patient.

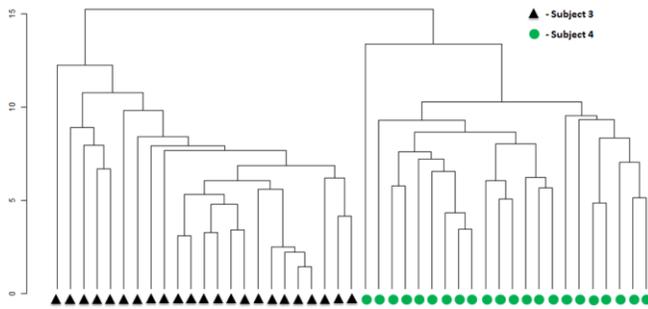

**Figure 12: Dendrogram produced by clustering days of two CH patients: Subject 3 has only "attack free" days, while Subject 4 is experiencing many CH attacks.**

While these results are preliminary, they show a promise that the sleep/wake patterns can be potentially used as personalized digital markers to create more effective therapeutic regimes and improve quality of life of millions of patients suffering from serious medical conditions.